\documentclass[a4paper]{article}
\usepackage{INTERSPEECH2022}

\title{Bottleneck Low-rank Transformers for Low-resource Spoken Language Understanding}
\name{Pu Wang, Hugo Van hamme}
\address{Department of Electrical Engineering-ESAT, KU Leuven, Belgium}
\email{pu.wang@esat.kuleuven.be, hugo.vanhamme@esat.kuleuven.be}

\begin{document}

\maketitle
\begin{abstract}
  End-to-end spoken language understanding (SLU) systems benefit from pretraining on large corpora, followed by fine-tuning on application-specific data. 
  The resulting models are too large for on-edge applications. For instance, BERT-based systems contain over 110M parameters.
  Observing the model is over-parameterized, we propose lean transformer structure where the dimension of the attention mechanism is automatically reduced using group sparsity. 
  We propose a variant where the learned attention subspace is transferred to an attention bottleneck layer.
  In a low-resource setting and without pre-training, the resulting compact SLU model achieves accuracies competitive with pre-trained large models.
\end{abstract}

\section{Introduction}

Spoken language understanding (SLU) systems that infer users' intents speech draw increased attention since the adoption of voice assistants such as Apple Siri and Amazon Alexa \cite{Ballati2018Siri, coucke2018snips, kumar2017just}. 
A traditional SLU system is the pipeline of an automatic speech recognition (ASR) component which decodes transcriptions from input speech and a natural language understanding (NLU) component which extracts meaning (intents) from the decoded transcriptions. To address several drawbacks of the pipeline structure such as error propagation, more and more research moves to end-to-end (E2E) SLU systems that directly map speech to intents. 

An E2E system is usually built with complex neural networks for sequence modeling and requires massive amounts of training data to outperform the traditional cascaded ASR-NLU systems.
However, in-domain intent-labeled speech data is scarce compared with ASR and NLU corpora.
Therefore, an emerging research interest is joint training or pre-training an E2E SLU system with a variety of related tasks including ASR \cite{1, 2, 3, 4, 6}, NLU \cite{1, 3, 4}, masked LM prediction \cite{1, 4}, and hypothesis prediction from text \cite{1, 2}. The most successful designs are transformer-based large-scale models \cite{6} including the pre-trained BERT model \cite{7, 8} and the default E2E ASR configuration in ESPnet \cite{4}. These models are huge and rely on large computational resources. For instance, the multilingual BERT-base model has 110M parameters and the GPT-2 model has 117M parameters \cite{9}. They are typically too large to store on an edge device thus requiring a cloud server, raising connectivity and privacy concerns. Moreover, large-scale models cause over parameterization issues when fine-tuning for low-resourced SLU tasks. This motivates work on compact SLU systems that run on the edge device itself. 

One traditional way to reduce model size involves factorized matrix representations \cite{9, 10, 13, 14, 15, 16, 17, 19, 21, 22}. Chen et al. \cite{11} apply this idea to linear projection layers in transformers. Akin to factorized TDNN models with the bottleneck layers, one linear projection layer with $m\times n$ weight matrix can be replaced with two stacked linear layers: $A$ of shape of $m \times r$ stacked on $B$ of shape $r\times n$, with rank $r \ll m,n$. This way, the number of parameters reduces from $n\times m$ to $r\times(n+m)$. Further research on low-rank approximations in transformer attention weight matrices chooses a fixed value for the rank of the factorization, i.e. the dimension of the bottleneck layer $r$, e.g. \cite{14, 22}, which led to performances degradation. 
Others utilize matrix factorization (MF) with knowledge distillations (KD). Mao et al. \cite{15} and Saghir et al. \cite{19}
firstly pre-train a (Distil)BERT model and then compress all linear layers in the (Distil)BERT model by using singular value decomposition (SVD) and pick the top $r$ singular values as the factorized weights ($r$ therefore is the chosen rank of the bottleneck layer). The weights are further fine-tuned on the target tasks. However, Chen et al. \cite{12} proved that the weight matrices in transformer linear layers are not low rank, i.e. solely pre-training of a (Distil)BERT model will not always achieve a low-rank representation. 
Moreover, Panahi et al. \cite{9} show that errors from stacking $r$-rank factorized linear transformations into a deep network add up which constrains the outputs by the smallest of ranks of the decomposed layers.

In this paper, we focus on designing a low-rank transformer structure whose dimensions are inferred by introducing a rank-related penalty term into the training. The well-trained model intrinsically holds a low-rank property without degradation of performance. The low-rank design is applied to the light transformer \cite{23} in a SLU context with very scarce training data. The light transformer is a compact version of the vanilla transformer \cite{24}. \cite{23} shows a successful implementation of the compact low-resource SLU system built with the light transformer combined with the capsule network which has only 1.3M parameters but achieves 98.8\% accuracy on FluentSpeech Command data without pre-training. 

The light transformer is firstly trained on the task-specific dataset with a rank-related penalty to Key and Query to get the low-rank self-attention representations. After that, a bottleneck transformer is built by inserting a bottleneck linear projection layer for Keys and Queries in the original multi-head attention layer. The bottleneck layer and its dimension are determined from the low-rank self-attention representations of a well-trained low-rank light transformer. The bottleneck transformer is then retrained using the same task specific dataset. 
The proposed structure is tested on the public SLU tasks comparing with advanced larger-scale pre-training models including the BERT and ESPnet. The contribution of this paper is hence: (1) Presenting a successful approach for low-rank approximation in transformer attention without performance loss in low-resource SLU (2) Proposing a compact low-rank (bottleneck) transformer-based SLU structure trained only on the application-specific data, which has competitive performances with advanced pre-training models. 

In section 2, we explain the light transformer and rank-related penalty firstly. The bottleneck transformer and the overall structure of the designed SLU system are introduced in this section as well. Section 3 discusses the specific experimental setting for evaluation, and the corresponding results will be presented in section 4. In section 5 we will conclude our work.
\section{Model}
\subsection{The baseline light transformer}
The light transformer is a light version of the vanilla transformer which includes a low-dimensional (light) relative position encoding (PE) matrix \cite{23}. We will shortly recap vanilla transformers first \cite{24}. The vanilla transformer is a layer-stacked model with each layer containing one multi-head self-attention block and one feed-forward block. The output of the multi-head attention is a linear projection of the concatenated multiple scaled dot-product self-attention operations as shown in Eq.~\ref{eq1}
\begin{equation}
  MultiHead = W_{out}Concat(Attn^1,Attn^2,...Attn^N)
  \label{eq1}
\end{equation}

In each attention head, the feature representation of the sequential data is first linearly transformed into a sequence of Keys, Values and Queries. The feature representation in the next layer is built as a non-linear mapping of a weighted average of the Values. The weight of each Value is determined by the similarity between the Key and the Query, as measured by a dot-product (Eq.~\ref{eq2}). 
\begin{equation}
  Attn^n = \sum_j \text{softmax}(S_{ij}) Vx_j \text{ with } S_{ij} = \frac{x_j^TK^TQx_i}{\sqrt{d_k}}
  \label{eq2}
\end{equation}
where $x$ is the feature representation or so-called content embedding. $Q, K, V\in\mathbb{R}^{d_k \times d}$ are weight matrices of the Query, the Key, and the Value respectively. $d$ is the dimension of the content embedding $x$. $d_k$ is the dimension of the transformer model. Softmax sums to unity over $j$.

To account for order information, the vanilla transformer will add a  $d$-dimensional absolute PE to the content embedding $x$ which requires the network to learn in which subspace relevant data variation occurs and in which subspace position is represented. To avoid the hassle brought by the additive high-dimensional PE, we introduced 6-dimensional relative position encoding in \cite{23}. It is defined by Eq.~\ref{eq3}.
\begin{equation}
   p_t^T=\begin{bmatrix}
   \cos\frac{2\pi t}{L} \,
   \sin\frac{2\pi t}{L} \,
   \cos\frac{2\pi t}{M_1} \,
   \sin\frac{2\pi t}{M_1} \,
   \cos\frac{2\pi t}{M_2} \,
   \sin\frac{2\pi t}{M_2}
   \end{bmatrix}
   \label{eq3}
\end{equation}
where $t$ is the position, $L=100$ to encode global sentence-level position,
$M_1$ and $M_2$ are small integers which are normally chosen as 4 and 8 to provide sufficient temporal resolution at phone and word level (at a frame shift of 40~ms).

Compared with the absolute PE applied in the vanilla transformer, the 6-d PE is concatenated to the content embedding which pre-defines a content-related and a location-related weight matrices $Q$ and $K$.
\begin{equation}
    Q=\begin{bmatrix}
    Q_c & 0 \\
    0 & Q_p
    \end{bmatrix}
    K=\begin{bmatrix}
    K_c & 0 \\
    0 & K_p
    \end{bmatrix}
    \label{eq4}
\end{equation}
where $Q_c$  and $K_c$ are content-related weight matrices, $Q_p$ and $K_p$ are position-related weights. The attention output is therefore presented by Eq.~\ref{eq5}.
\begin{equation}
    S_{ij} = \frac{x_j^T K_c^T Q_c x_i}{\sqrt{d_k}}+\frac{p_j^T K_p^T Q_p p_i}{\sqrt{d_p}}
    \label{eq5}  
\end{equation}

To consider arbitrary relative relations between Query and Key, we further replace the absolute 6-d $p$ with the relative PE $p_{i-j}$ as shown in Eq.~\ref{eq6}. $u \in \mathbb{R}^{d_p}$ is a trainable location-based parameter.
\begin{equation}
    S_{ij} = \frac{x_j^T K_c^T Q_c x_i}{\sqrt{d_k}}+\frac{p_{i-j}^Tu}{\sqrt{d_p}}
    \label{eq6}  
\end{equation}
\subsection{Low-rank light transformer}
\begin{figure}[t]
    \centering
    \includegraphics[width=0.54\linewidth,trim={0 0 0 10},clip]{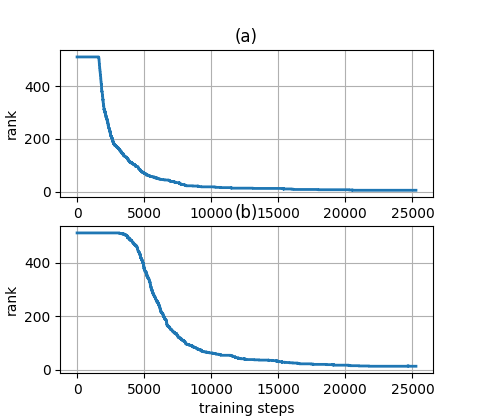}\includegraphics[width=0.46\linewidth,trim={210 0 210 0},clip]{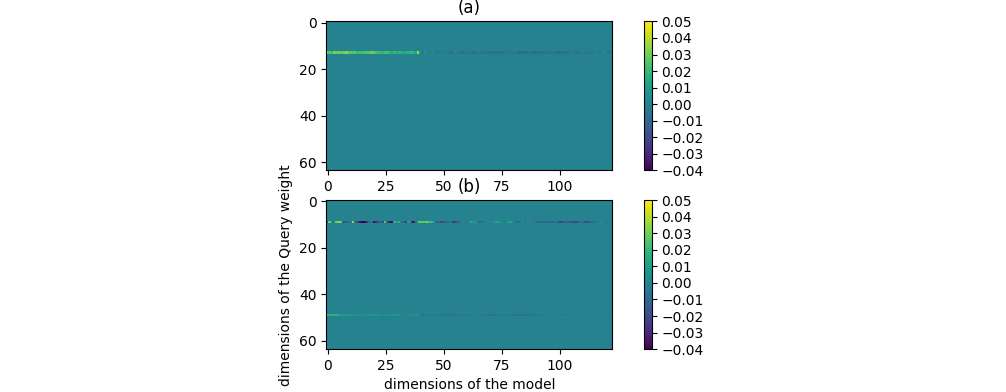}
    \caption{The dynamic of ranks and values of the $Q_c$ matrix training with the penalty scale (a) 0.0005, (b) 0.0001.}
    \label{fig1}
\end{figure}
To achieve a low-rank attention model, group sparse regularization terms for $Q_c$ and $K_c$ are added to the total loss during training:
\begin{equation}
    L_{gs} = \lambda_{gs} \sum_{\text{layers}}\sum_{\text{heads}}\sum_{k=1}^{d_k} \left( \sqrt{\sum_{d=1}^d Q{_{c_{k,d}}}^2} + \sqrt{\sum_{d=1}^d K{_{c_{k,d}}}^2} \right)
\label{eq8}
\end{equation}


The group sparsity penalty will prefer to steer complete rows in $Q_c$ and $K_c$ to zero, thus reducing the rank of these matrices. Figure~\ref{fig1} shows how the rank decreases during training (left) and the resulting sparsity in $Q_c$ (right) with different penalty scales $\lambda_{gs}$.

Normalisation with $\sqrt{d_k}$ in Eq. \ref{eq2} is introduced in \cite{24} to avoid gradient scale problems. When training with $L_{gs}$, the effective inner product dimension decreases during training. We therefore replace the scale parameter $\sqrt{d_k}$ of Eq.~\ref{eq6} with the dynamic rank of the weight matrices. The \textit{total} rank of $Q_c$ ($K_c$) is defined in Eq.~\ref{eq9}. Here we only show the expression of $Q_c$ since the ranks of these two matrices are always identical. Indeed, suppose $Q_c$ has a zero row, any non-zero entry in $K_c$ will not affect the transformer behavior, while the regularization term can be reduced by setting its value to zero. The roles of $Q_c$ and $K_c$ can be swapped in this argument.
\begin{equation}
    r_{sum}=\sum_{\text{heads}}\sum_{k=1}^{d_k}\left\{
    \begin{array}{cc}
    1 \text{ if } \sum_{d=1}^{d}\left|{Q_c}_{k,d}\right|\geq 0.001&  \\
    0 \text{ if } \sum_{d=1}^{d}\left|{Q_c}_{k,d}\right|< 0.001   & 
    \end{array}
\right.
\label{eq9}
\end{equation}
Notice that the individual ranks of the different heads and layers will be different if it leads to a lower loss. It is hence different from tuning the $d_k$ parameter to a lower value.
\subsection{Bottleneck light transformer}
\begin{figure}[t]
    \centering
   \includegraphics[width=1\linewidth,trim={65 0 10 63},clip]{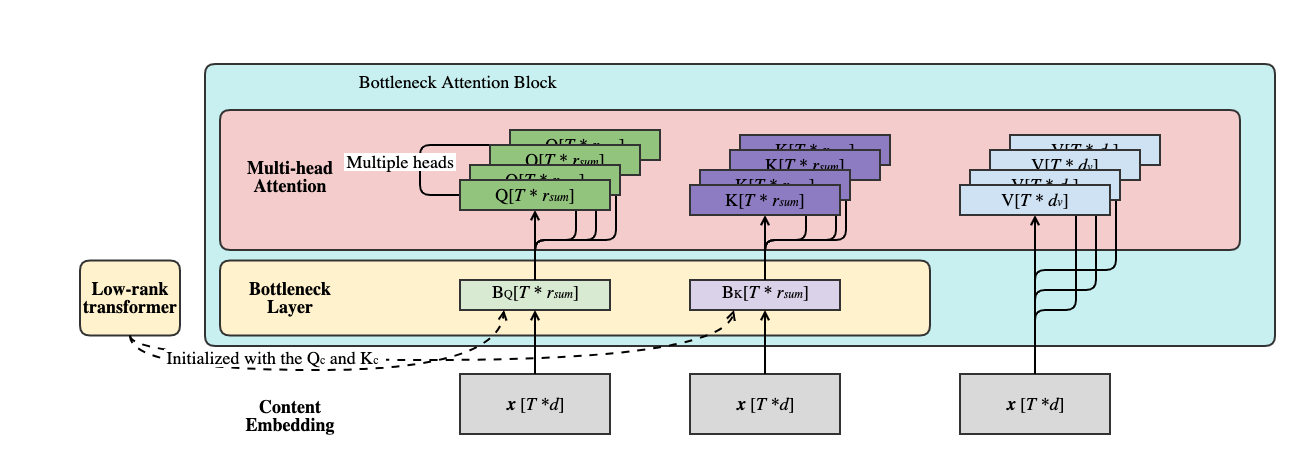}
    \caption{Bottleneck transformer: bottleneck layer applies to Keys and Values.}
    \label{fig2}
\end{figure}
The heads in a transformer attend to different data properties. 
The low-rank light transformer results in very sparse attention matrices with only a few or no dimensions per head left.
The idea of this section is to give the transformer a second opportunity to learn combinations of properties it finds useful in attention. We therefore introduce a bottleneck layer in the light transformer spanning the subspace deemed relevant by the group-sparse heads in the low-rank transformer as shown in the Figure~\ref{fig2}. 
The bottleneck light transformer introduces a different bottleneck layer of dimension $r_{sum}$ for Query and Key before the multi-head attention layer. The bottleneck is however common to all heads. 
The weight matrices of the bottleneck layer $B_Q$ and $B_K$ are initialized with the low-rank $Q_c$ and $K_c$ transferred from the low-rank light transformer respectively. We do not introduce a bottleneck layer for the Values: the dimension of each head for Values is unchanged from the (low-rank) light transformer.
For example, if a low-rank transformer starts with 8 heads with 64-dimensions per head and ends up with 8 heads with on average 2-dimension per head, the dimension of the bottleneck layer will be 16. Query and Key will still have 8 heads but with 16-dimension per head and Value will keep its 64-dimension per head. The bottleneck transformer is re-trained with the same dataset as the low-rank transformer.

\subsection{Compact SLU decoder}
The whole structure of the compact SLU system is 
built from \cite{23} with the encoder-decoder concept. The filterbank (Fbank) inputs are first processed by the low-rank (bottleneck) transformer encoder to yield the high-level representations. The high-level representations are fed into a 2-layer capsule network decoder yields the task information. There are 32 hidden capsules with 64 dimensions in the primary capsule layer and one output capsule for each output slot labels with 16-dimensions in the output capsule layer. The detailed description of capsule decoder can be found on \cite{31}. 
The low-rank (bottleneck) transformer consists of 3 layers. Every layer has an 8-head parallel attention layer and a feed-forward layer with 2048 hidden nodes. The dimension of the Value of each attention head is 64, the dimensions of Key and Query of the low-rank transformer in each attention head are started from 64. The dimensions of Key and Query of the bottleneck low-rank transformer are initialized by the well-trained low-rank transformer. To further speed up the training, self-attentions were restricted to considering only a neighborhood of size 5 centered around the respective output position. 
To reduce sequence length, we use two 2-dimensional convolution layers with kernel size $(3, 3)$ at the very beginning to implement 4-fold down-sampling in time.

\section{Experiments}

\subsection{Dataset}
The low-rank (bottleneck) transformer is applied to two public SLU corpora. 

Domotica addresses 27 home automation tasks for Dutch pathological speech. Notice that the pretraining approach is difficult here, for there is a lack of disordered speech data. This database is collected from 17 speakers in three time phases \cite{29}. We use the subsets named Domotica 3 and Domotica 4 which contains 4180 utterances in total.

FluentSpeech Commands (FSC) records 31 home automation tasks from 97 English speakers. This dataset is challenging with varied command phrasings which contains 248 different phrasing with 30043 utterances in total \cite{30}. Pretaining is possible here.

\subsection{Experimental setup}
To simulate the low-resourced scenarios, we extract small fraction of samples from each dataset to form the training set. All experiments are conducted under a speaker-independent setting using
10-fold cross-validation.
Specifically, for the Domotica data, we randomly select 2 samples for each tasks from each speaker as training samples, which totals 735 samples (because not all speakers perform all tasks). The remaining 3440 samples are used for testing. We evaluate the performances by the F1 score of the detected slot values.

The FSC data is divided into train set (23132 utterances), validation set (3118 utterances) and test set (3793) utterances \cite{25}. We randomly select 10\%, 30\%, 60\% and 100\% data from the train set, and report intent accuracy \cite{23} on the test set. The accuracy metric is defined as the accuracy of all slots for an utterance taken together \cite{25}.



\subsection{Hyper parameters}
The proposed compact SLU system is trained with the Adam optimizer with $\beta_1=0.9,\beta_2=0.98$, $\epsilon=10^{-16}$ and warmup \cite{24}. The final model is constructed by averaging the model parameters of the last 10 training steps. To regularize during training, we apply dropout with a rate of $P_{drop=0.1}$ to each sub-layer, including the content and position embeddings.
\section{Results}
\subsection{Domotica dataset}
\begin{figure}[t]
    \centering
    \includegraphics[width=1\linewidth,trim={5 0 25 10},clip]{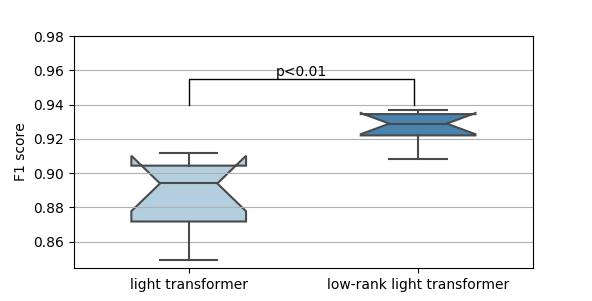}
    \caption{F1 scores of light transformer and low-rank transformer (low rank with $\lambda_{gs}=0.0005$, $r \approx 4$)}
    \label{fig4}
\end{figure}

We first compare slot value F1 results of the low-rank light transformer with the light transformer. The low-rank transformer is trained with a penalty scale of $0.0005$, the ranks of well-trained self-attention weight matrices are $4$. Figure~\ref{fig4} shows the box plot of results of the $10$ times experiments with the Wilcoxon significance test of these two models. The p-value is shown in the Figure~\ref{fig4} as well. 

With an average rank of $4$ in self-attention, the low-rank transformer significantly outperforms the rank-64 light transformer, showing an efficient assignment of learnable parameters for scarce training data. One could also argue that the transformer model simply benefits from regularization in an over-parameterized model. 
We therefore train light transformer models with a smaller common dimension $d_k$ for Query and Key to test if performance can be preserved by using a more compact structure.
Secondly, we apply standard L2 regularization as well as (unstructured) sparsity-inducing L1 regularization.

\begin{figure}[t]
    \centering
    \includegraphics[width=1\linewidth,trim={10 0 30 10},clip]{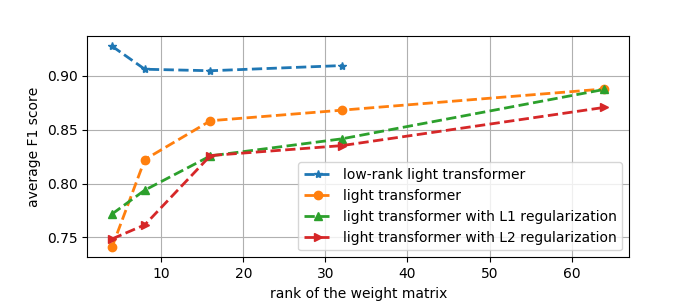}
    \caption{Average F1 scores of low-rank transformer and light transformer with different ranks}
    \label{fig6}
\end{figure}

We 
investigate whether the performance gains come from the group sparsity or regularization in Figure~\ref{fig6}, which shows the average F1 scores of various light transformers at different rank choices for the weight matrices. The performance of the transformer dramatically degrades with smaller dimension of Query and Key, which indicates that the benefits brought by the low-rank penalty are not easily obtained by tuning $d_k$. Introducing L1 or L2 regularization also does not improve performance, which shows the light transformer structure does benefit from group sparsity. Hence, we conclude that group sparsity effectively finds the subspaces that are important for the attention heads.

We compare the average F1 results with other state-of-art encoders including two advanced large-scale pre-trained models \cite{4, 26}. The performance and model sizes are summarized in Table ~\ref{tab2}. RCCN is a GRU-based SLU system proposed in \cite{31}. ESPnet is a transformer-based model evaluated in \cite{4} with a 12-layer encoder and a 6-layer decoder. To adapt to the Dutch Domotica data, we pre-train this model accompanied with CTC loss using Dutch Copas disordered speech \cite{33}. Kaldi is a TDNN-F-based model proposed in \cite{32}, pre-trained on the Dutch Copas data as well. All compared models are combined with the same capsule network decoder for intent classification as described in the Section.2.4.
\begin{table}[tbh]
  \caption{Average F1 scores and model scales of Domotica.}
  \label{tab2}
  \centering
  \begin{tabular}{ccc}
    \toprule
    Method & F1 score & \# of param. \\
    \midrule
    Light transformer \cite{23}    & $0.888$ & $1.3M$            \\
           + low-rank         & $0.930$ & $1.3M$               \\
           + bottleneck   & $0.950$ &  $1.1M$              \\
    \hline\hline
    RCCN \cite{31} & $0.908$ & $2.5M$ \\
    ESPnet (pre-training) \cite{4} & $0.931$ & $27M+0.7M$ \\
    Kaldi (pre-training) \cite{32} & $0.939$ & $19M+0.7M$ \\
    \bottomrule
  \end{tabular}
\end{table}

From Table~\ref{tab2}, the light transformer model slightly worse than GRU-based RCCN model. After introducing the low-rank penalty, it outperforms the RCCN model by 3\% absolute which indicates the light transformer gives too much freedom to self-attention mechanism that can be constrained with low-rank penalty. In general, low-rank light transformer get comparable results with two pre-training models shows that the low-rank light transformer is capable to extract expressive representations from dysarthric speech inputs. After remedy from inserting bottleneck layer, it outperforms the two pre-training model. One possible reason is speaker-independent dysarthric SLU task is hard to benefit from pre-training while bottleneck transformer is benefits from transfer learning of low-rank transformer. 
\subsection{FSC dataset}
Following the experimental setting in \cite{23}, we simulate an insufficient training situation by randomly choosing 10\%, 30\%, 60\%, and 100\% data from the train set. We summarize the accuracy results of proposed SLU system as well as mainstream approaches including RNN-based, CNN-based, vanilla transformer-based and advanced pre-training models in the Table~\ref{tab3}. 

Without pre-training, the low-rank (bottleneck) transformer outperforms most other approaches. Although the bottleneck transformer is slightly worse than the evaluated pre-trained models with 10\% training data, with the full training samples, the bottleneck transformer (without pre-training) still shows competitive results with other large-scale pre-trained models.

\begin{table}[tbh]
  \caption{Average accuracy of FSC.}
  \label{tab3}
  \centering
  \setlength{\tabcolsep}{1.4mm}{
  \begin{tabular}{ccccc}
    \toprule
    Method & $10\%$  & $30\%$ & $60\%$  & $100\%$ \\
    \midrule
    Light transformer \cite{23}    & $91.8$ & $96.7$ & $98.3$ & $98.9$            \\
           + low-rank         & $93.8$ & $97.7$ & $98.7$ & $99.2$               \\
           + bottleneck   & $96.9$ & $98.4$ & $98.9$ & $99.3$                \\
    \hline\hline
    RNN \cite{28} & $-$ & $-$ & $-$ & $95.3$ \\
    RNN + capsule network \cite{23} & $81.5$ & $94.9$ & $96.7$ & $98.1$ \\
    Transformer-based \cite{27} & $-$ & $-$ & $-$ & $97.6$ \\
    \hline
    RNN + SincNet \cite{27}& $-$ & $-$ & $-$ & $96.1$ \\
    + pre-training \cite{27}& $-$ & $-$ & $-$ & $97.2$ \\
    \hline
    BERT-based (pre-training)\cite{7} & $-$ & $-$ & $-$ &
    $99.1$ \\
    ESPnet-based (pre-training) \cite{4} & $-$ & $-$ & $-$ &
    $99.6$ \\
    CNN-based (pre-training)\cite{26} & $-$ & $-$ & $-$ & $99.3$ \\
    \hline
    Lugosh et al. \cite{25} & $88.9$ & $-$ & $-$ & $96.6$ \\
    +AM+pre-training \cite{25} & $97.9$ & $-$ & $-$ & $98.9$ \\
    \hline
    Self-attention + BLSTM \cite{6} & $-$ & $-$ & $-$ & $99.1$ \\
    + pre-training \cite{6} & $-$ & $-$ & $-$ & $99.3$ \\
    \bottomrule
  \end{tabular}}
\end{table}

\section{Conclusions}
Transformer models are over-parameterized for low-resource SLU applications. 
We therefore proposed using group sparsity to automatically infer a low-rank attention model. The learned attention subspace of the low-rank transformer is then transferred to an attention bottleneck layer to form a compact transformer variant.
The compact SLU system built on the (bottleneck) low-rank transformer achieved $95\%$ F1 score on the Domotica disordered speech data and $99.3\%$ accuracy on the FSC data without pre-training, which is comparable to advanced approaches with pre-training ($93.9\%$ F1 score on Domotica and the best $99.6\%$ accuracy on FSC). 

\section{Acknowledgements}
The research was supported by the program of China Scholarship Council No.$201906090275$ and the Flemish Government under “Onderzoeksprogramma AI Vlaanderen”.

\bibliographystyle{IEEEtran}
\bibliography{mybib}

\end{document}